%% file: egpaper_for_review.tex
\documentclass[10pt,twocolumn,letterpaper]{article}

\usepackage{cvpr}
\usepackage{times}
\usepackage{epsfig}
\usepackage{graphicx}
\usepackage{amsmath}
\usepackage{amssymb}
\usepackage[table,dvipsnames]{xcolor}
\usepackage{multirow}

\usepackage[pagebackref=true,breaklinks=true,letterpaper=true,colorlinks,bookmarks=false]{hyperref}

\def\ijcbPaperID{43} %

\begin{document}

\title{Identity Overlap Between Face Recognition Train/Test Data:\\ Causing Optimistic Bias in Accuracy Measurement}

\author{Haiyu Wu$^{1}$, Sicong Tian$^{2}$, Jacob Gutierrez$^{1}$, Aman Bhatta$^{1}$\\ Kağan Öztürk$^1$, Kevin W. Bowyer$^{1}$\\
$^{1}$University of Notre Dame\\ $^2$Indiana University South Bend\\}

\maketitle
\thispagestyle{empty}

\input{sec/0_abstract}    
\input{sec/1_intro_v2}
\input{sec/2_related_work}
\input{sec/3_identity_overlap}

\input{sec/4_fr_method_comparison}
\input{sec/5_experiments}
\input{sec/6_conclusion}
{\small
\bibliographystyle{ieee}
\bibliography{egbib}
}

\end{document}

%% file: sec/0_abstract.tex
\begin{abstract}

A fundamental tenet of pattern recognition is that overlap between training and testing sets causes an optimistic accuracy estimate.
Deep CNNs for face recognition are trained for N-way classification of the identities in the training set.
Accuracy is commonly estimated as average 10-fold classification accuracy on image pairs from test sets such as LFW, CALFW, CPLFW, CFP-FP and AgeDB-30.
Because train and test sets have been independently assembled, images and identities in any given test set may also be present in any given training set.
In particular, our experiments reveal a surprising degree of identity and image overlap between the LFW family of test sets and the MS1MV2 training set. 
Our experiments also reveal identity label noise in MS1MV2.
We compare accuracy achieved with same-size MS1MV2 subsets that are identity-disjoint and not identity-disjoint with LFW, to reveal the size of the optimistic bias.
Using more challenging test sets from the LFW family, we find that the size of the optimistic bias is larger for more challenging test sets.
Our results highlight the lack of and the need for identity-disjoint train and test methodology in  face recognition research.
\end{abstract}

%% file: sec/1_intro_v2.tex
\section{Introduction}
\label{sec:intro}
\input{figures/teaser-figure}

Two main factors have powered remarkable advances in face recognition:
1) more advanced training methods~\cite{normface, magface, arcface, circleloss, adaface, transface, uniface}, and 2) creation of larger and cleaner training datasets~\cite{casia-webface, vggface, vggface2, ms-celeb-1m, arcface, ms1mv3, webface260m, glint360k}.
However, after CosFace~\cite{cosface}, accuracy improvements have hit a plateau.
For the five commonly used test sets – LFW~\cite{lfw}, CFP-FP~\cite{cfpfp}, AgeDB-30~\cite{agedb-30}, CALFW~\cite{calfw}, CPLFW~\cite{cplfw} –  reported average accuracy improvement since CosFace is less than 1\%.
Also, comparison of smaller accuracy differences is problematic because methods often vary in backbone implementations, image augmentations, evaluation preprocessing, and other details.
More reliable comparison of algorithms can be made if all the training and evaluation details are the same.
But a potentially more important issue that has gone  `under the radar' concerns the popular web-scraped, in-the-wild training and test sets.
Training sets and test sets have been created as independent efforts by different groups, which means that identities and images in any given test set may also be in any given training set.
This raises important unanswered questions.
\textbf{\textit{How great is the overlap of identities and images between commonly-used train and test sets?}}
\textbf{\textit{How great is the effect of overlap between the test sets and training sets on the estimated accuracy of algorithms?}}
\textbf{\textit{Do some algorithms exploit the train/test overlap to achieve higher accuracy by ``memorizing'' identities in the training data?}}

Contributions of this paper include:
\begin{itemize}
\item Demonstrating that train/test overlap of identities and images does exist, by reverse-engineering the overlap between MS1MV2 and LFW.
\item Providing a uniform training environment that enables more controlled accuracy comparison between algorithms.
\item Exploring the impact of identity overlap on a range of tests of varying difficulty.
\end{itemize}

%% file: figures/teaser-figure.tex
\begin{figure}
    \centering
    \begin{subfigure}[b]{1\linewidth}
    \captionsetup[subfigure]{labelformat=empty}
        \includegraphics[width=\linewidth]{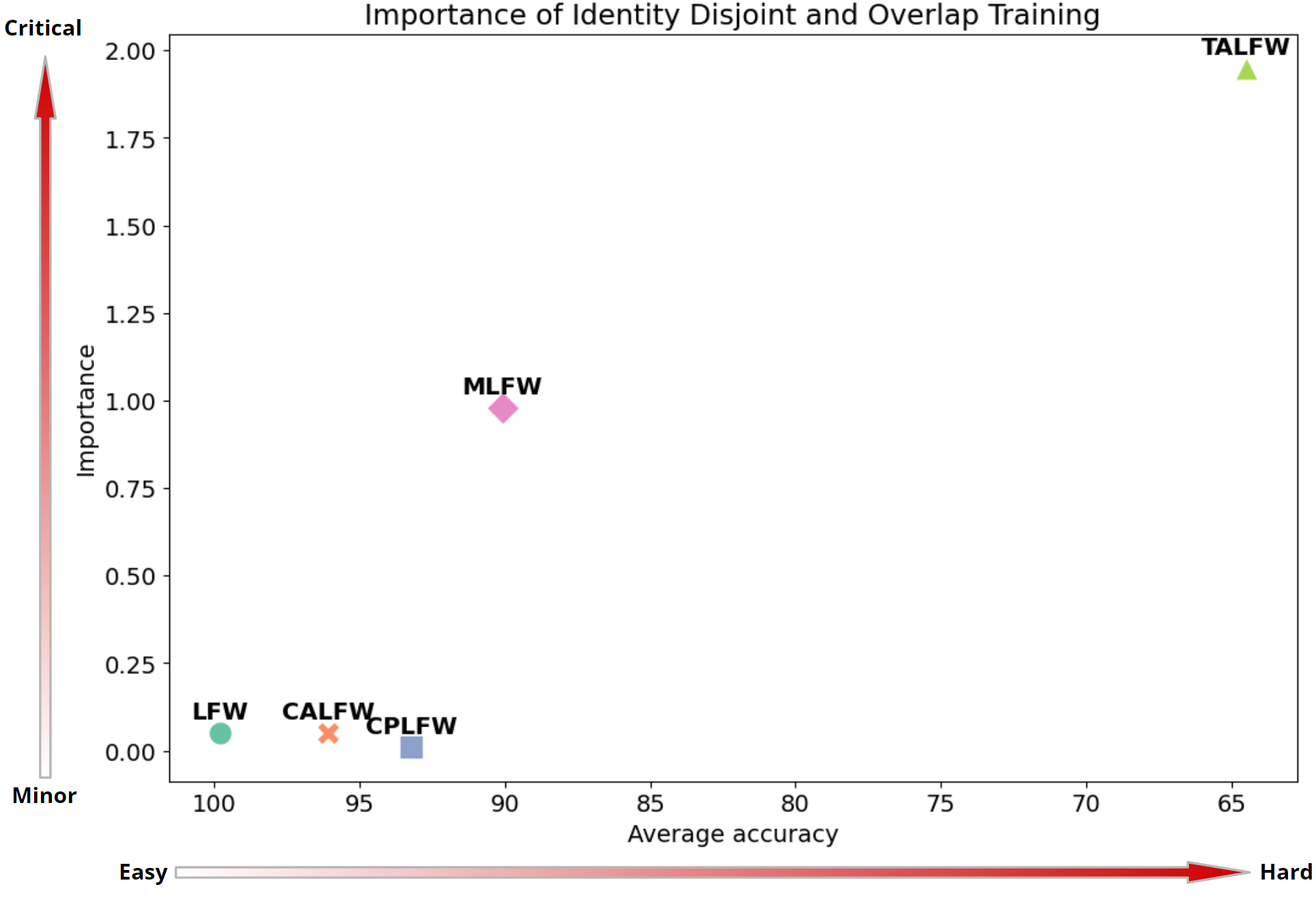}
    \end{subfigure}
   \caption{Importance of identity-disjoint training in face recognition. It involves five test sets, which have same identities, for different face recognition tasks. On the vertical axis, "importance" of identity-disjoint training is the accuracy difference between identity-overlapped and identity-disjoint training. On the horizontal axis, "difficulty" of the test set is the average accuracy value of two training approaches.}
\label{fig:teaser-figure}
\end{figure}

%% file: sec/2_related_work.tex
\section{Related Work}
\label{sec:related}
\input{figures/manual_results}
\paragraph{Face Recognition Methods.}
An evolution of face recognition loss functions began with NormFace~\cite{normface}, which transformed the softmax function by normalizing the magnitudes of feature vector $X$ and weight vector $W$ to 1. This change altered the network's output from $XW^{T}$ to a simple $cos\theta$. Following this,  techniques were explored to incorporate a margin to enhance feature discriminability. CosFace~\cite{cosface} introduced a  margin to intra-class features, expressed as $cos\theta - m$. SphereFace~\cite{sphereface} proposed a proportional increase in the cosine value, represented by $mcos\theta$. ArcFace~\cite{arcface} moved to angular space, adding a margin as $cos(\theta + m)$ and $cos(m_{1}\theta + m_{2}) - m_{3}$. 
The concept of adaptive margins emerged as a alternative. CurricularFace~\cite{curricularface} incorporated curricular learning principles into the loss function. MagFace~\cite{magface} adjusted the margin based on recognizability. AdaFace~\cite{adaface} leveraged the relationship between feature norm and image quality to dynamically modify the margin. Additional research focused on varying backbones~\cite{groupface, transface} and other aspects of the loss function~\cite{uniface}.

\paragraph{Face Recognition Training Sets.}
To boost face recognition research, several large-scale face datasets have been introduced. CASIA-WebFace~\cite{casia-webface} comprises 0.5M images from 10K identities, with the dataset being cleaned through semi-automated methods. VGGFace2~\cite{vggface2} is an expanded and more refined version of VGGFace~\cite{vggface}, containing 3.3M images from 9K identities. This dataset underwent cleaning by employing a classifier to identify and remove outliers within each identity. MS-Celeb-1M~\cite{ms-celeb-1m} utilized a combination of entity keys from a knowledge base and personal names to help ensure that web-scraped images correspond to the targeted identities. This dataset contained 10M images from 100K celebrities. Cleaner versions of this dataset, MS1MV2~\cite{arcface} and MS1MV3~\cite{ms1mv3}, were developed using different techniques. Additionally, MS1MV2 was used to create Glint360K~\cite{glint360k}, which features 17M  images from 360K identities. The current largest dataset is WebFace260M~\cite{webface260m}, boasting 260M images from 4M identities. Three subsets of WebFace260M were assembled: WebFace4M, WebFace12M, and WebFace42M.

\paragraph{Face Recognition Testing Sets.}
LFW~\cite{lfw} contains 13,233 images of 5,749 individuals and established a standard paradigm for evaluating face recognition. For evaluation, the dataset includes 6K image pairs, split between genuine  (same person) pairs and impostor (different people) pairs. These pairs are divided into 10 folds, each containing 300 genuine and 300 impostor pairs. A model's accuracy is estimated by the average accuracy across this 10-Fold cross-validation. Subsequent test sets, such as YTF~\cite{ytf}, 
adopted this evaluation format. 
Additionally,  factors such as age~\cite{agedb-30, calfw} and pose~\cite{cfpfp, cplfw} are emphasized by specific additional test sets.
In a different approach, sets like IJB-A, IJB-B, and IJB-C~\cite{ijba, ijbb, ijbc}, released by IARPA, evaluate models based on verification, identification, and image vs. video performance, reporting accuracy as True Positive Rates (TPRs) at various False Positive Rates (FPRs). There are also specific test sets designed to assess bias in face recognition, including BFW~\cite{bfw}, DemogPairs~\cite{demogpairs}, and BA-test~\cite{ba-test}.

\paragraph{Effects of Duplicate and Near-duplicate images.}
A recent analysis of face recognition datasets \cite{duplicateimages} examines the impact of duplicate  and near-duplicate images within a particular dataset.
They analyze LFW, TinyFace, Adience, CASIA-WebFace, and a cleaned MS-Celeb-1M variant.
They find that removing duplicate images from within training identities has a relatively  minor effect on estimated accuracy of the trained model.
Existence of duplicate and near-duplicate images in the Celeb-A dataset was previously analyzed by \cite{Wu_2023}.

\paragraph{Difference of our work.}
Our work is fundamentally different from other work mentioned above in that our work looks at identity and image overlap {\it across} a training and a  testing set, and how this causes an optimistic bias in estimated accuracy.
We are not aware of any previous work in face recognition that analyzes identity overlap across train and test sets.

%% file: figures/manual_results.tex
\begin{figure*}[t]
    \centering
    \captionsetup[subfigure]{labelformat=empty}
    \begin{subfigure}[b]{1\linewidth}
        \begin{subfigure}[b]{1\linewidth}
            \begin{subfigure}[b]{0.48\linewidth}
                \begin{subfigure}[b]{0.24\linewidth}
                    \includegraphics[width=\linewidth]{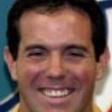}
                    \caption{Brian\_Griese\_1}
                \end{subfigure}
                \begin{subfigure}[b]{0.24\linewidth}
                    \includegraphics[width=\linewidth]{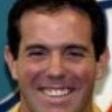}
                    \caption{0\_5870303/40}
                \end{subfigure}
                \begin{subfigure}[b]{0.24\linewidth}
                    \includegraphics[width=\linewidth]{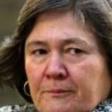}
                    \caption{Clare\_Short\_4}
                \end{subfigure}
                \begin{subfigure}[b]{0.24\linewidth}
                    \includegraphics[width=\linewidth]{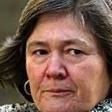}
                    \caption{0\_5836550/78}
                \end{subfigure}
            \end{subfigure}
            \begin{subfigure}[b]{0.48\linewidth}
                \begin{subfigure}[b]{0.24\linewidth}
                    \includegraphics[width=\linewidth]{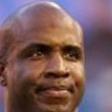}
                    \caption{Barry\_Bonds\_1}
                \end{subfigure}
                \begin{subfigure}[b]{0.24\linewidth}
                    \includegraphics[width=\linewidth]{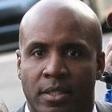}
                    \caption{0\_5836919/74}
                \end{subfigure}
                \begin{subfigure}[b]{0.24\linewidth}
                    \includegraphics[width=\linewidth]{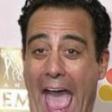}
                    \caption{Brad\_Garrett\_4}
                \end{subfigure}
                \begin{subfigure}[b]{0.24\linewidth}
                    \includegraphics[width=\linewidth]{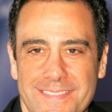}
                    \caption{0\_5840877/112}
                \end{subfigure}
            \end{subfigure}
        \end{subfigure}

        \begin{subfigure}[b]{1\linewidth}
            \begin{subfigure}[b]{0.48\linewidth}
                \begin{subfigure}[b]{0.24\linewidth}
                    \includegraphics[width=\linewidth]{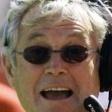}
                    \caption{Dick\_Vermeil\_1}
                \end{subfigure}
                \begin{subfigure}[b]{0.24\linewidth}
                    \includegraphics[width=\linewidth]{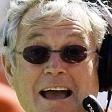}
                    \caption{0\_5885817/7}
                \end{subfigure}
                \begin{subfigure}[b]{0.24\linewidth}
                    \includegraphics[width=\linewidth]{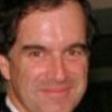}
                    \caption{Garry\_Trudeau\_1}
                \end{subfigure}
                \begin{subfigure}[b]{0.24\linewidth}
                    \includegraphics[width=\linewidth]{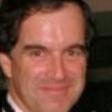}
                    \caption{0\_5856109/65}
                \end{subfigure}
            \caption{(a) Same Image}
            \end{subfigure}            
            \begin{subfigure}[b]{0.48\linewidth}
                \begin{subfigure}[b]{0.24\linewidth}
                    \includegraphics[width=\linewidth]{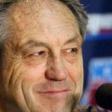}
                    \caption{Eddie\_Sutton\_1}
                \end{subfigure}
                \begin{subfigure}[b]{0.24\linewidth}
                    \includegraphics[width=\linewidth]{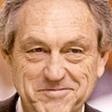}
                    \caption{0\_5867491/15}
                \end{subfigure}
                \begin{subfigure}[b]{0.24\linewidth}
                    \includegraphics[width=\linewidth]{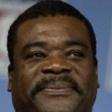}
                    \caption{Eddie\_Murray\_1}
                \end{subfigure}
                \begin{subfigure}[b]{0.24\linewidth}
                    \includegraphics[width=\linewidth]{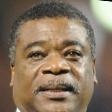}
                    \caption{0\_5848896/50}
                \end{subfigure}
            \caption{(b) Same Identity}
            \end{subfigure}
        \end{subfigure}
    \end{subfigure}
   \caption{Examples of the same identities and the same images occur in both LFW and MS1MV2. More examples are shown in Figures 4-7 of Supplementary Material.
   }
\label{fig:identity-overlap}
\end{figure*}

%% file: sec/3_identity_overlap.tex
\section{Identity Overlap Detection}
\label{sec:id-overlap}
\input{figures/special-cases}
\input{figures/lfw-ms1mv2-violin}

Training a deep CNN face matcher is an N-way classification problem, with the goal of getting all images of each of the N identities classified into the right ``bin’’.
Standard testing of a face matcher is a 2-way image-pair classification problem, with the goal of maximizing the correct classification as genuine and impostor pairs.
Face recognition research has come to use training and test sets of web-scraped images, assembled at different times by different groups.
This situation suggests that identities and images in any given test set could easily also be in any given training set.
In this section, we reverse engineer the overlap between the archetypal test set, LFW, and training set, MS1MV2. Note that multiple test sets emphasizing different dimensions of difficulty (pose, age, ...) are based on the same identities as in LFW, and so in this sense LFW represents a family of related datasets.

For each of the 12,000 images in the LFW test set, we used automated matching to select the two most similar images from MS1MV2, resulting in similarity scores for 24,000 image pairs.
Matching was done with ArcFace-R100 trained with Glint360K\footnote{https://github.com/deepinsight/insightface/tree/master/model\_zoo\#list-of-models-by-various-depth-iresnet-and-training-datasets} to extract the features
and cosine similarity for matching.
The left violin plot in Figure~\ref{fig:lfw-ms1mv2-violin} shows the distribution of the 24,000 similarity values.
There is a bulge below 0.5, which could naturally correspond identities in LFW that are not also in MS1MV2.
The bulge in the range of 0.6 to 0.9 could then correspond to identities in LFW that are also in MS1MV2, and the minor bulge above 0.9 correspond to images in LFW that have a near-duplicate in MS1MV2.

To check this interpretation of the distribution of similarity scores, human annotators marked the 24,000 image pairs as same-identity or different-identity.
For an image pair marked as same-identity, annotators also marked duplicate images, or not.
Results confirm that there are 384 (5\%) LFW images with a duplicate or near-duplicate image in MS1MV2, and 2,009 (46.93\%) identities in LFW that are also in MS1MV2.
The fraction of duplicate images detected across the two datasets is roughly in line with what might be expected based on previous work analyzing fraction of duplicate images within a single dataset \cite{duplicateimages,Wu_2023}.
But the fraction of overlapped identities across the two datasets, which has not been studied before, seems surprisingly high.
Example image pairs across the two datasets are shown in Figure~\ref{fig:identity-overlap}.

\input{table/performance}
The middle and right violin plots in Figure~\ref{fig:lfw-ms1mv2-violin} show the distribution of similarity scores for image pairs marked as same identity and different identity, respectively.
Even though a distinct boundary at 0.5 exists in the left violin plot, there are some high-similarity image pairs that a human rater judged as not the same identity, and some low-similarity pairs that a human rater judged as the same identity.
We re-examined 183 image pairs that had similarity $\geq$0.8 from the matcher but were marked as not same identity by the human rater, and 331 image pairs with similarity $\leq$0.4 but marked as not the same identity by the human rater.
Figure~\ref{fig:special-cases} shows sample image pairs for these cases, where each image pair, we believe, is from the same identity. 
While the face matcher similarity seems robust to factors of age and expression, it seems vulnerable to varied brightness level and to occlusion.
And, human rater labeling as same / different identity also has some errors.

In the next section, we use our knowledge of the identity overlap between LFW and MS1MV2 to explore how identity overlap between train and test data affects accuracy estimates.
Note that the cross-pose LFW and the cross-age LFW datasets are based on the same identities as the original LFW, and so we have the identity overlap of all three of these test sets with MS1MV2.

%% file: figures/special-cases.tex
\begin{figure*}[t]
    \centering
    \begin{subfigure}[b]{1\linewidth}
    \captionsetup[subfigure]{labelformat=empty}
        \begin{subfigure}[b]{1\linewidth}
            \begin{subfigure}[b]{0.48\linewidth}
                \begin{subfigure}[b]{0.24\linewidth}
                    \includegraphics[width=\linewidth]{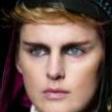}
                \end{subfigure}
                \begin{subfigure}[b]{0.24\linewidth}
                    \includegraphics[width=\linewidth]{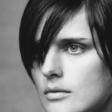}
                \end{subfigure}
                \begin{subfigure}[b]{0.24\linewidth}
                    \includegraphics[width=\linewidth]{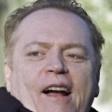}
                \end{subfigure}
                \begin{subfigure}[b]{0.24\linewidth}
                    \includegraphics[width=\linewidth]{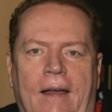}
                \end{subfigure}
            \end{subfigure}
            \begin{subfigure}[b]{0.48\linewidth}
                \begin{subfigure}[b]{0.24\linewidth}
                    \includegraphics[width=\linewidth]{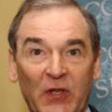}
                \end{subfigure}
                \begin{subfigure}[b]{0.24\linewidth}
                    \includegraphics[width=\linewidth]{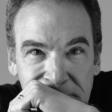}
                \end{subfigure}
                \begin{subfigure}[b]{0.24\linewidth}
                    \includegraphics[width=\linewidth]{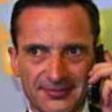}
                \end{subfigure}
                \begin{subfigure}[b]{0.24\linewidth}
                    \includegraphics[width=\linewidth]{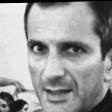}
                \end{subfigure}
            \end{subfigure}
        \end{subfigure}

        \begin{subfigure}[b]{1\linewidth}
            \begin{subfigure}[b]{0.48\linewidth}
                \begin{subfigure}[b]{0.24\linewidth}
                    \includegraphics[width=\linewidth]{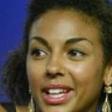}
                \end{subfigure}
                \begin{subfigure}[b]{0.24\linewidth}
                    \includegraphics[width=\linewidth]{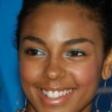}
                \end{subfigure}
                \begin{subfigure}[b]{0.24\linewidth}
                    \includegraphics[width=\linewidth]{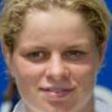}
                \end{subfigure}
                \begin{subfigure}[b]{0.24\linewidth}
                    \includegraphics[width=\linewidth]{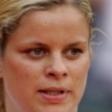}
                \end{subfigure}
            \caption{HSNS: high similarity measure, marked not same person}
            \end{subfigure}            
            \begin{subfigure}[b]{0.48\linewidth}
                \begin{subfigure}[b]{0.24\linewidth}
                    \includegraphics[width=\linewidth]{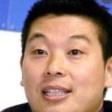}
                \end{subfigure}
                \begin{subfigure}[b]{0.24\linewidth}
                    \includegraphics[width=\linewidth]{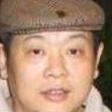}
                \end{subfigure}
                \begin{subfigure}[b]{0.24\linewidth}
                    \includegraphics[width=\linewidth]{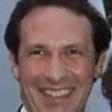}
                \end{subfigure}
                \begin{subfigure}[b]{0.24\linewidth}
                    \includegraphics[width=\linewidth]{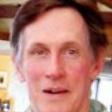}
                \end{subfigure}
            \caption{LSTS: low similarity measure, marked same person}
            \end{subfigure}
        \end{subfigure}
    \end{subfigure}
   \caption{Examples of the high similarity but marked as not the same identity (HSNS) and the low similarity but marked as the same identity (LSTS).}
\label{fig:special-cases}
\end{figure*}

%% file: figures/lfw-ms1mv2-violin.tex
\begin{figure}
    \centering
    \begin{subfigure}[b]{1\linewidth}
    \captionsetup[subfigure]{labelformat=empty}
        \includegraphics[width=\linewidth]{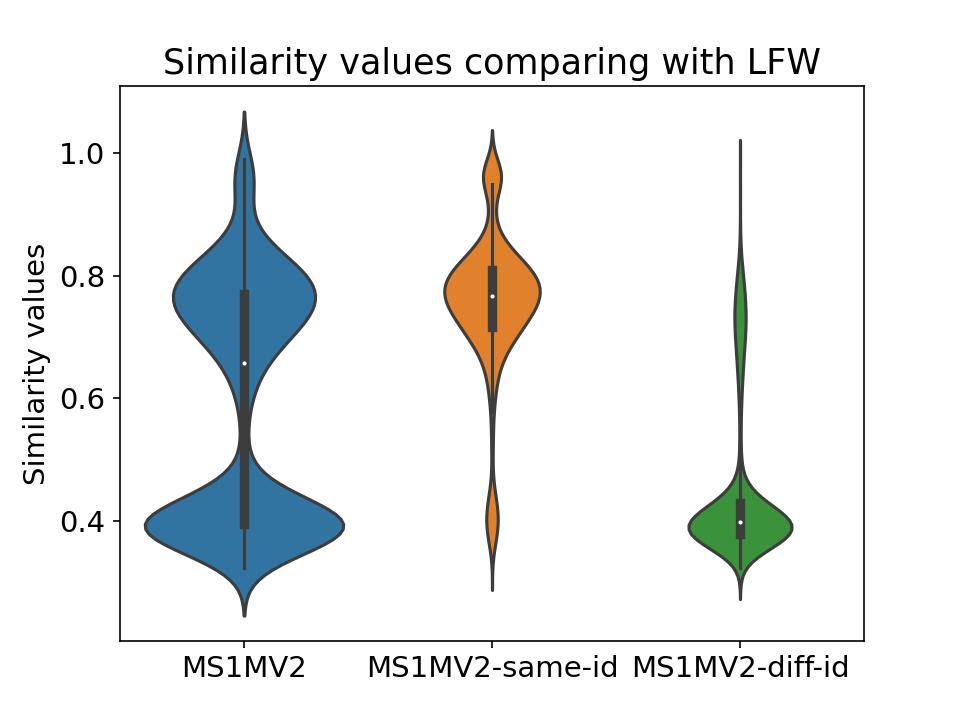}
    \end{subfigure}
   \caption{Similarity distribution of top two matches in MS1MV2. Based on the manual marking result, the similarity distribution of the pairs marked as same identity and different identities are at middle and right.}
\label{fig:lfw-ms1mv2-violin}
\end{figure}

%% file: table/performance.tex
\begin{table*}
    \centering
    \begin{tabular}{|c|c|c|c|c|c|c|c|}
    \hline
        Methods & Type & LFW & CFP-FP & CPLFW & AgeDB-30 & CALFW & AVG\\ \hline
        CosFace (m=0.35)\cite{cosface} & Reported & - & - & - & - & - & -\\ \hline
        SphereFace (m=1.7)\cite{sphereface} & Reported & - & - & - & - & - & -\\ \hline
        ArcFace (m=0.5)\cite{arcface} & Reported & 99.82 & - & 92.08 & - & 95.45 & -\\ \hline
        CurricularFace (m=0.5)\cite{curricularface} & Reported & 99.80 & 98.37 & 93.13 & \textcolor[HTML]{1A09F3}{\textbf{98.32}} & \textcolor[HTML]{1A09F3}{\textbf{96.20}} & 97.16\\ \hline
        MagFace\cite{magface} & Reported & \textcolor[HTML]{1A09F3}{\textbf{99.83}} & 98.46 & 92.87 & 98.17 & 96.15 & 97.10\\ \hline
        AdaFace\cite{adaface} & Reported & 99.82 & \textcolor[HTML]{1A09F3}{\textbf{98.49}} & \textcolor[HTML]{1A09F3}{\textbf{93.53}} & 98.05 & 96.08 & 97.19\\ \hline \hline
        
        CosFace (m=0.35) & Retrained-O & 99.82 & \textcolor[HTML]{1A09F3}{\textbf{98.43}} & 92.85 & \textcolor[HTML]{1A09F3}{\textbf{98.18}} & 95.93 & 97.04\\ \hline
        SphereFace (m=1.7) & Retrained-O & \textcolor[HTML]{1A09F3}{\textbf{99.83}} & 98.07 & 92.73 & \textcolor[HTML]{1A09F3}{\textbf{98.18}} & \textcolor[HTML]{1A09F3}{\textbf{96.17}} & 97.00\\ \hline
        ArcFace (m=0.5) & Retrained-O & 99.78 & 98.17 & 93.07 & 98.00 & 96.05 & 97.03\\ \hline
        CurricularFace (m=0.5) & Retrained-O & 99.80 & 98.13 & 92.57 & 97.90 & 96.07& 96.89\\ \hline
        MagFace & Retrained-O & \textcolor[HTML]{1A09F3}{\textbf{99.83}} & 98.13 & 92.33 & 98.12 & 96.07 & 96.90\\ \hline
        AdaFace & Retrained-O & 99.82 & 98.21 & \textcolor[HTML]{1A09F3}{\textbf{93.12}} & 98.12 & \textcolor[HTML]{1A09F3}{\textbf{96.17}} & 97.09\\ \hline \hline

        CosFace (m=0.35) & Retrained-OF & 99.75 & 98.41 & 93.25 & \textcolor[HTML]{1A09F3}{\textbf{98.33}} & 95.93 & 97.14\\ \hline
        SphereFace (m=1.7) & Retrained-OF & \textcolor[HTML]{1A09F3}{\textbf{99.83}} & 98.40 & 92.67 & 98.22 & 96.07 & 97.04\\ \hline
        ArcFace (m=0.5) & Retrained-OF & 99.80 & 98.49 & \textcolor[HTML]{1A09F3}{\textbf{93.35}} & 98.00 & 96.05 & 97.14\\ \hline
        CurricularFace (m=0.5) & Retrained-OF & 99.78 & 98.44 & 92.95 & 98.05 & 96.08 & 97.06\\ \hline
        MagFace & Retrained-OF & 99.82 & 98.21 & 92.67 & 98.15 & 96.13 & 97.00\\ \hline
        AdaFace & Retrained-OF & 99.82 & \textcolor[HTML]{1A09F3}{\textbf{98.63}} & 93.05 & 98.20 & \textcolor[HTML]{1A09F3}{\textbf{96.15}} & 97.17\\ \hline
    \end{tabular}
    \caption{Accuracy on benchmark datasets of SOTA methods trained with ResNet100 and MS1MV2. 
    Top block (type = reported) is  accuracy  reported in the cited paper; middle block (type = retrained-o) is accuracy of model retrained with original images for test; bottom block ((type = retrained-of) is retrained model with summed features of the original and flipped images for test.}
    \label{tab:test-performance}
\end{table*}

%% file: sec/4_fr_method_comparison.tex
\section{Face Recognition Methods Comparison}
\label{sec:fr-comparison}
\input{figures/one-matches-many}
Even though %
various methods in the literature may all use ResNet50 or ResNet100 backbones, the actual implementations are slightly different, in ways that can affect accuracy estimates. For data augmentation during training, random horizontal flip is used by all methods, but AdaFace also uses color jitter, random crop, and low resolution. 
Such a difference in augmentation methods can cause an observed accuracy difference that may incorrectly be attributed to algorithm difference rather than augmentation difference.
MagFace and AdaFace use BGR channel instead of RGB, and MagFace normalizes the images with \{{\tt mean=0., std=1.}\} rather than \{{\tt mean=0.5, std=0.5}\} to train and test the model. For evaluation, CosFace, SphereFace, ArcFace, CurricularFace, and AdaFace report average accuracy using the summation of the features of the original and the horizontal flip of the image, but MagFace reports the average accuracy with  features of only the original images. Moreover, MagFace uses clipped cosine distance rather than Euclidean distance to find the best threshold in validation. These various differences make ``apples to apples'' comparison of accuracy difficult.
We select three constant-margin methods (CosFace, SphereFace, ArcFace) and three adaptive margin methods (CurricularFace, MagFace, AdaFace) to compare, using the same implementation details for all six.
Our main point in this experiment is to assess how identity-disjoint train and test data impacts the estimated accuracy of matchers, and not to rank the six re-trained matchers.

We use a variant of ResNet100, specifically adapted for ArcFace training, as our primary backbone\footnote{https://github.com/deepinsight/insightface/}. To expedite the training, the Partial-FC~\cite{glint360k} strategy was applied across all six training methodologies. 
MS1MV2 is the training set. Images, formatted in RGB, were resized to 112x112. Each image underwent normalization \{mean=0.5, std=0.5\} and data augmentation through random horizontal flipping. Default hyper-parameter configurations were adopted as per the respective open-source code packages. SGD\cite{sgd} served as the optimizer, configured with a learning rate of 0.1, momentum of 0.9, and weight decay set at 0.0005. Training was conducted with a batch size of 256 over 20 epochs. For performance evaluation, we report accuracy metrics using solely the original images (Retrained-O) and a combination of original plus horizontally flipped images' features (Retrained-OF), to highlight any differences observed.

Table~\ref{tab:test-performance} shows the accuracy values of the reported and retrained models. The retrained models we used have the best performance on LFW, CFP-FP, and AgeDB-30 during training. The results show that summing the feature vectors of the original images and horizontally flipped images consistently improves average accuracy compared to using only the original feature vectors; the largest improvement is 0.11\% in average accuracy. 
Since summing the feature vectors of the original and flipped images consistently gives higher accuracy, the rest of our experiments and analyses use this approach. 
The  matchers' accuracies are competitive, with the difference between the highest and lowest average accuracy being 0.17\%, and no one method consistently surpasses the others across the five test sets.

%% file: figures/one-matches-many.tex
\begin{figure*}[t]
    \centering
    \begin{subfigure}[b]{0.8\linewidth}
    \captionsetup[subfigure]{labelformat=empty}
        \begin{subfigure}[b]{1\linewidth}
            \begin{subfigure}[b]{0.19\linewidth}
                \includegraphics[width=\linewidth]{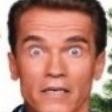}
            \end{subfigure}
            \begin{subfigure}[b]{0.19\linewidth}
                \includegraphics[width=\linewidth]{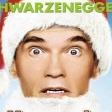}
            \end{subfigure}
            \begin{subfigure}[b]{0.19\linewidth}
                \includegraphics[width=\linewidth]{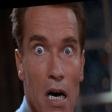}
            \end{subfigure}
            \begin{subfigure}[b]{0.19\linewidth}
                \includegraphics[width=\linewidth]{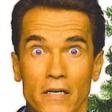}
            \end{subfigure}
            \begin{subfigure}[b]{0.19\linewidth}
                \includegraphics[width=\linewidth]{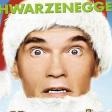}
            \end{subfigure}
        \end{subfigure}
        \begin{subfigure}[b]{1\linewidth}
            \begin{subfigure}[b]{0.19\linewidth}
                \includegraphics[width=\linewidth]{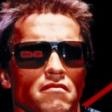}
            \end{subfigure}
            \begin{subfigure}[b]{0.19\linewidth}
                \includegraphics[width=\linewidth]{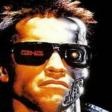}
            \end{subfigure}
            \begin{subfigure}[b]{0.19\linewidth}
                \includegraphics[width=\linewidth]{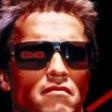}
            \end{subfigure}
            \begin{subfigure}[b]{0.19\linewidth}
                \includegraphics[width=\linewidth]{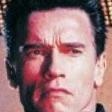}
            \end{subfigure}
            \begin{subfigure}[b]{0.19\linewidth}
                \includegraphics[width=\linewidth]{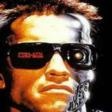}
            \end{subfigure}
        \end{subfigure}
        \begin{subfigure}[b]{1\linewidth}
            \begin{subfigure}[b]{0.19\linewidth}
                \includegraphics[width=\linewidth]{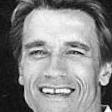}
            \end{subfigure}
            \begin{subfigure}[b]{0.19\linewidth}
                \includegraphics[width=\linewidth]{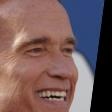}
            \end{subfigure}
            \begin{subfigure}[b]{0.19\linewidth}
                \includegraphics[width=\linewidth]{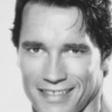}
            \end{subfigure}
            \begin{subfigure}[b]{0.19\linewidth}
                \includegraphics[width=\linewidth]{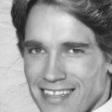}
            \end{subfigure}
            \begin{subfigure}[b]{0.19\linewidth}
                \includegraphics[width=\linewidth]{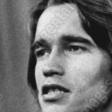}
            \end{subfigure}
            \label{fig:arnold}
        \end{subfigure}
        \end{subfigure}

   \caption{Examples of identity mis-categorization in MS1MV2. Top row: images from MS1MV2 identity folder 0\_5873183; middle row: images from MS1MV2 folder 0\_5876541; bottom row: images from MS1MV2 folder 0\_5898354. The three ``identities'' are all Arnold Schwarzenagger.}
\label{fig:one-match-many}
\end{figure*}

%% file: sec/5_experiments.tex
\section{Experiments}
\label{sec:id-overlap}
\input{table/or-dis-over}
\input{table/highest-lfws-performance}

\input{table/lfw-datasets}

The main purpose of this section is to explore how large of an optimistic bias is created in the face recognition accuracy estimate by identity overlap between the train and test sets.  However, the process of reverse engineering the identity overlap between LFW and MS1MV2 also uncovered a degree of identity labeling errors in MS1MV2, and so we first assess the impact of cleaning the identity labeling errors, and then assess the impact of identity overlap between train and test sets.

The experiments described in the previous section traced 2,009 identities in the LFW test set back to 2,195 identities in MS1MV2.
The ``extra’’ 186 identities could in principle be due to MS1MV2 either (a) having identity folders with one ore more incorrect images in them, such as a Danny DeVito folder containing an image of Arnold Schwarzeneggar, or (b) having images of one identity being split across multiple folders, such as Arnold Schwarzeneggar being split across folders according to characters in different movies.
We manually reviewed the instances of multiple identity mappings, and found that most were due to one real identity being split across multiple MS1MV2 identity folders.
Example images from three MS1MV2 folders that are all Arnold Schwarzeneggar are shown in Figure~\ref{fig:one-match-many}.  Additional examples of this kind of identity labeling problem are shown in the Supplementary Material.

Based on the LFW identities mapping to the 2,195 identities in MS1MV2, we created three subsets of MS1MV2.
The ID-Disjoint subset drops the 2,195 identity folders that overlap with LFW identities.
The ID-Overlap-R (for ``raw’’) subset keeps the identity folders that overlap with LFW identities, and drops the same number (2,195) of randomly-selected non-identity-overlapped folders.
The ID-Overlap-C (for ``cleaned’’) subset is the same as ID-Overlap-R, but with the identities in MS1MV2 that are split across multiple folders each merged into one folder.
We then trained each of the six face matchers with each of the three versions of the MS1MV2 training set.
The resulting accuracies are shown in Table~\ref{tab:test-or-dis-over}.
The AVG-LFWs column is the average across the three LFW family test sets, which share the same set of identities.
We present this separately because we know the identity overlap of LFW and MS1MV2 but do not yet know the identity overlap of CFP-FP and AgeDB-30 with MS1MV2.

Note that, for each of the six models, the ID-Overlap-C training achieves higher AVG-LFWs accuracy than either the ID-Disjoint training or the ID-Overlap-R training.
The ID-Overlap-C training achieving higher accuracy than the ID-Overlap-R training shows that even a relatively small fraction of the total identities being split across multiple folders represents a serious confounding factor to training an accurate model.
Put simply, cleaner training data is always better.

The ID-Overlap-C training achieving higher accuracy than the ID-Disjoint training shows that a cleaned training set that has identity overlap with the test set results in an optimistic accuracy bias.
Put simply, identity overlap between train and test data causes a ``too high’’ accuracy estimate.

Table~\ref{table:best-lfws} compares the average performance on LFW, CPLFW, CALFW, between the model trained with the vanilla MS1MV2 and LFW-identity-cleaned MS1MV2. It shows that, even though the vanilla MS1MV2 has more training samples (i.e. 150K images from 2,195 + 186 = 2,381 identities), the model trained with ID-Overlap-C can perform better with the benefit of a cleaned identity-overlap version.

Therefore, if the identities exist in both train and test, even though the images are not the same and the task is easy (i.e. $\geq99.75$\% on LFW across six methods), the model trained with such training sets still achieves a higher estimated accuracy than the model trained with identity-disjoint training sets.  Fundamental machine learning knowledge principles tell us that same image should not occur in both training and testing sets. 
In the context of face recognition more specifically, this section proves that identities plays a similar role as images in other contexts.

\subsection{Importance of Identity Overlap and Disjoint}

The identity overlap existing between training and testing sets gives the model a chance to ``cheat'', and achieve a consistently higher estimated accuracy. However, it seems like the performance difference is not significantly higher than when it does not ``cheat'' since 
the largest average accuracy difference is 0.13\% on SphereFace and the largest accuracy difference between these two is 0.32\% on CPLFW. 
\textit{Is it because the face recognition methods are too powerful to fail at recognition in general? Or is the task  not challenging enough?}. To answer this question, we conducted another experiment using two more challenging datasets - Masked LFW (MLFW)~\cite{mlfw} and Transferable Adversarial LFW (TALFW)~\cite{talfw}.
MLFW contains the same images in CALFW but with automatically added synthetic masks. It is built for face recognition task during the COVID-19 pandemic. TALFW uses the same images in LFW but with automatically added transferable adversarial noise. Both datasets contain the same identities in LFW but have much lower accuracy.

\input{figures/importance}
Table~\ref{tab:lfws-performance} shows the performances of the methods trained with ID-Overlap-C and ID-Disjoint versions. To better understand the relationship between dataset difficulty and identity overlap importance, we measure the difficulty level and importance level with respect to the accuracy value and accuracy difference on each test set. For example, LFW is the easiest dataset (highest accuracy) and TALFW is the hardest dataset (lowest accuracy). As for importance, for CosFace, the importance of identity overlap on LFW is 0.03\% and 0.95\% on TALFW. Figure~\ref{fig:importance-figure} indicates that identity overlap between training and testing dataset causes a larger difference on more challenging datasets. On the most difficult test set, TALFW, the smallest performance difference is 0.34\% and the largest performance difference is 1.95\%. Both values are significantly higher than the improvement between the reproduced highest and lowest accuracy among face recognition methods. Therefore, avoiding identity overlap between training and testing sets is necessary to better evaluate model performance. We also notice that face recognition methods perform worse on CPLFW than on CALFW, but the importance value on CPLFW is not always lower than on CALFW. Our speculation is that the horizontal-flip augmentation used in training somewhat increases the robustness of the methods to pose problems. Interestingly, there are not-a-face images in the CPLFW dataset (distributed by InsightFace\footnote{https://github.com/deepinsight/insightface/tree/master/recognition/\_datasets\_}), which could explain the low accuracy on this dataset. Details are in the Supplementary Material.

%% file: table/or-dis-over.tex
\begin{table*}
    \centering
    \begin{tabular}{|c|c|c|c|c|c|c|c|}
    \hline
        Methods & Type & LFW & CFP-FP & CPLFW & AgeDB-30 & CALFW & AVG-LFWs\\ \hline
        CosFace (m=0.35) & ID-Overlap-R & \cellcolor{gray!25} \textbf{99.82} & 98.43 & \cellcolor{gray!25} 93.05 & 97.98 & \cellcolor{gray!25} 96.00 & 96.29\\ \hline
        CosFace (m=0.35) & ID-Disjoint & \cellcolor{gray!25} 99.75 & 98.24 & \cellcolor{gray!25} 93.20 & 98.12 & \cellcolor{gray!25} 96.07 & 96.34\\ \hline 
        CosFace (m=0.35) & ID-Overlap-C & \cellcolor{gray!25} 99.78 & \textbf{98.61} & \cellcolor{gray!25} \textbf{93.27} & \textbf{98.30} & \cellcolor{gray!25} \textbf{96.18} & \textcolor[HTML]{e60000}{\textbf{96.41}}\\ \hline\hline
        SphereFace (m=1.7) & ID-Overlap-R & \cellcolor{gray!25} \textbf{99.83} & 98.20 & \cellcolor{gray!25} 92.78 & \textbf{98.23} & \cellcolor{gray!25} \textbf{96.08} & 96.23\\ \hline
        SphereFace (m=1.7) & ID-Disjoint & \cellcolor{gray!25} 99.78 & \textbf{98.33} & \cellcolor{gray!25} 92.78 & 97.97 & \cellcolor{gray!25} 96.03 & 96.20\\ \hline
        SphereFace (m=1.7) & ID-Overlap-C & \cellcolor{gray!25} 99.82 & 98.31 & \cellcolor{gray!25} \textbf{93.10} & 97.98 & \cellcolor{gray!25} 96.07 & \textcolor[HTML]{e60000}{\textbf{96.33}}\\ \hline\hline
        ArcFace (m=0.5) & ID-Overlap-R & \cellcolor{gray!25} \textbf{99.83} & \textbf{98.44} & \cellcolor{gray!25} 92.90 & 98.05 & \cellcolor{gray!25} 96.05 & 96.26\\ \hline
        ArcFace (m=0.5) & ID-Disjoint & \cellcolor{gray!25} 99.80 & 98.33 & \cellcolor{gray!25} 92.90 & \textbf{98.25} & \cellcolor{gray!25} \textbf{96.15} & 96.28\\ \hline 
        ArcFace (m=0.5) & ID-Overlap-C & \cellcolor{gray!25} 99.72 & 98.41 & \cellcolor{gray!25} \textbf{93.12} & 97.97 & \cellcolor{gray!25} 96.10 & \textcolor[HTML]{e60000}{\textbf{96.31}}\\ \hline\hline
        CurricularFace (m=0.5) & ID-Overlap-R & \cellcolor{gray!25} \textbf{99.82} & 98.36 & \cellcolor{gray!25} 92.70 & 98.07 & \cellcolor{gray!25} 96.10 & 96.20\\ \hline
        CurricularFace (m=0.5) & ID-Disjoint & \cellcolor{gray!25} \textbf{99.82} & \textbf{98.44} & \cellcolor{gray!25} \textbf{92.87} & \textbf{98.22} & \cellcolor{gray!25} 96.05 & 96.25\\ \hline 
        CurricularFace (m=0.5) & ID-Overlap-C & \cellcolor{gray!25} 99.78 & \textbf{98.44} & \cellcolor{gray!25} 92.82 & 98.03 & \cellcolor{gray!25} \textbf{96.18} & \textcolor[HTML]{e60000}{\textbf{96.26}}\\ \hline\hline
        MagFace & ID-Overlap-R & \cellcolor{gray!25} \textbf{99.82} & 98.43 & \cellcolor{gray!25} 92.82 & 98.07 & \cellcolor{gray!25} \textbf{96.08} & 96.24\\ \hline
        MagFace & ID-Disjoint & \cellcolor{gray!25} 99.78 & 98.39 & \cellcolor{gray!25} 92.87 & \textbf{98.08} & \cellcolor{gray!25} 96.03 & 96.23\\ \hline 
        MagFace & ID-Overlap-C & \cellcolor{gray!25} 99.80 & \textbf{98.49} & \cellcolor{gray!25} \textbf{93.08} & \textbf{98.08} & \cellcolor{gray!25} 96.07 & \textcolor[HTML]{e60000}{\textbf{96.32}}\\ \hline\hline
        AdaFace & ID-Overlap-R & \cellcolor{gray!25} 99.78 & 98.30 & \cellcolor{gray!25} 93.15 & 98.20 & \cellcolor{gray!25} 96.05 & 96.33\\ \hline
        AdaFace & ID-Disjoint & \cellcolor{gray!25} 99.78 & \textbf{98.50} & \cellcolor{gray!25} 93.22 & \textbf{98.23} & \cellcolor{gray!25} 96.07 & 96.36\\ \hline
        AdaFace & ID-Overlap-C & \cellcolor{gray!25} \textbf{99.83} & 98.36 & \cellcolor{gray!25} \textbf{93.25} & 98.10 & \cellcolor{gray!25} \textbf{96.12} & \textcolor[HTML]{e60000}{\textbf{96.40}}\\ \hline
    \end{tabular}
    \caption{Performance of SOTA methods trained with three versions of MS1MV2. Columns in gray are the results on LFW and its variations.[R: Raw, C: Corrected]}
    \label{tab:test-or-dis-over}
\end{table*}

%% file: table/highest-lfws-performance.tex
\begin{table*}[]
\centering
\begin{tabular}{|c|c|c|c|c|c|c|c|}
\hline
    & Type         & CosFace & SphereFace & ArcFace & CurricularFace & MagFace & AdaFace \\ \hline
\multirow{2}{*}{AVG-LFWs} & Retrained-OF & 96.31   & 96.19      & \textbf{96.40}   & 96.25 & 96.21   & 96.34   \\ \cline{2-8}
 & ID-Overlap-C & \textbf{96.41}   & \textbf{96.33}      & 96.31   &   \textbf{96.26}   & \textbf{96.32}   & \textbf{96.40} \\ \hline
\end{tabular}
\caption{Performance comparison between SOTA methods trained with vanilla MS1MV2 (Retrain-OF) and trained with LFW identity corrected MS1MV2 (ID-Overlap-C) with less identities and images.}
\label{table:best-lfws}
\end{table*}

%% file: table/lfw-datasets.tex
\begin{table*}
    \centering
    \begin{tabular}{|c|c|c|c|c|c|c|c|c|}
    \hline        Methods & Type & LFW & CPLFW & CALFW & MLFW & TALFW & AVG\\ \hline
        CosFace (m=0.35) & ID-Overlap-C & \textbf{99.78} & \textbf{93.27} & \textbf{96.18} & \textbf{90.38} & \textbf{62.48} & \cellcolor{gray!25} \textbf{88.42}\\ \hline
        CosFace (m=0.35) & ID-Disjoint & 99.75 & 93.20 & 96.07 & 89.37 & 61.53 & \cellcolor{gray!25} 87.98\\ \hline \hline
        SphereFace (m=1.7) & ID-Overlap-C & \textbf{99.82} & \textbf{93.10} & \textbf{96.07} & \textbf{89.68} & \textbf{65.52} & \cellcolor{gray!25} \textbf{88.84}\\ \hline
        SphereFace (m=1.7) & ID-Disjoint & 99.78 & 92.78 & 96.03 & 89.53 & 65.18 & \cellcolor{gray!25} 88.66\\ \hline \hline
        ArcFace (m=0.5) & ID-Overlap-C & \textbf{99.72} & \textbf{93.12} & 96.10 & \textbf{90.45} & \textbf{67.02} & \cellcolor{gray!25} \textbf{89.28}\\ \hline
        ArcFace (m=0.5) & ID-Disjoint & 99.80 & 92.90 & \textbf{96.15} & 89.87 & 66.18 & \cellcolor{gray!25} 88.98\\ \hline \hline
        CurricularFace (m=0.5) & ID-Overlap-C & 99.78 & \textbf{92.83} & \textbf{96.18} & \textbf{90.12} & \textbf{63.83} & \cellcolor{gray!25} \textbf{88.55}\\ \hline
        CurricularFace (m=0.5) & ID-Disjoint & \textbf{99.82} & 92.80 & 95.97 & 89.48 & 62.82 & \cellcolor{gray!25} 88.18\\ \hline \hline
        MagFace & ID-Overlap-C & \textbf{99.80} & \textbf{93.08} & \textbf{96.07} & \textbf{89.85} & \textbf{62.23} & \cellcolor{gray!25} \textbf{88.21}\\ \hline
        MagFace & ID-Disjoint & 99.78 & 92.87 & 96.03 & 89.43 & 61.23 & \cellcolor{gray!25} 87.85\\ \hline \hline
        AdaFace & ID-Overlap-C & \textbf{99.83} & \textbf{93.23} & \textbf{96.12} & \textbf{90.58} & \textbf{65.48} & \cellcolor{gray!25} \textbf{89.05}\\ \hline
        AdaFace & ID-Disjoint & 99.78 & 93.22 & 96.07 & 89.60 & 63.53 & \cellcolor{gray!25} 88.44\\ \hline \hline
        AVG-FOR-ALL & ID-Overlap-C & \textbf{99.79} & \textbf{93.27} & \textbf{96.18} & \textbf{90.38} & \textbf{62.48} & \cellcolor{gray!25} \textbf{88.42}\\ \hline
        AVG-FOR-ALL & ID-Disjoint & 99.75 & 93.20 & 96.07 & 89.37 & 61.53 & \cellcolor{gray!25} 87.98\\ \hline \hline
        Optimistic Bias & ID-Overlap-C - ID-Disjoint & \textcolor[HTML]{e60000}{\textbf{0.04}} & \textcolor[HTML]{e60000}{\textbf{0.07}} & \textcolor[HTML]{e60000}{\textbf{0.11}} & \textcolor[HTML]{e60000}{\textbf{1.01}} & \textcolor[HTML]{e60000}{\textbf{0.95}} & \cellcolor{gray!25} \textcolor[HTML]{e60000}{\textbf{0.44}}\\ \hline
    \end{tabular}
    \caption{Performance of SOTA methods trained with two versions of MS1MV2 on five test sets with the same set of LFW identities.[C: Corrected]}
    \label{tab:lfws-performance}
\end{table*}

%% file: figures/importance.tex
\begin{figure*}[h]
    \centering
    \begin{subfigure}[b]{1\linewidth}
    \captionsetup[subfigure]{labelformat=empty}
        \includegraphics[width=\linewidth]{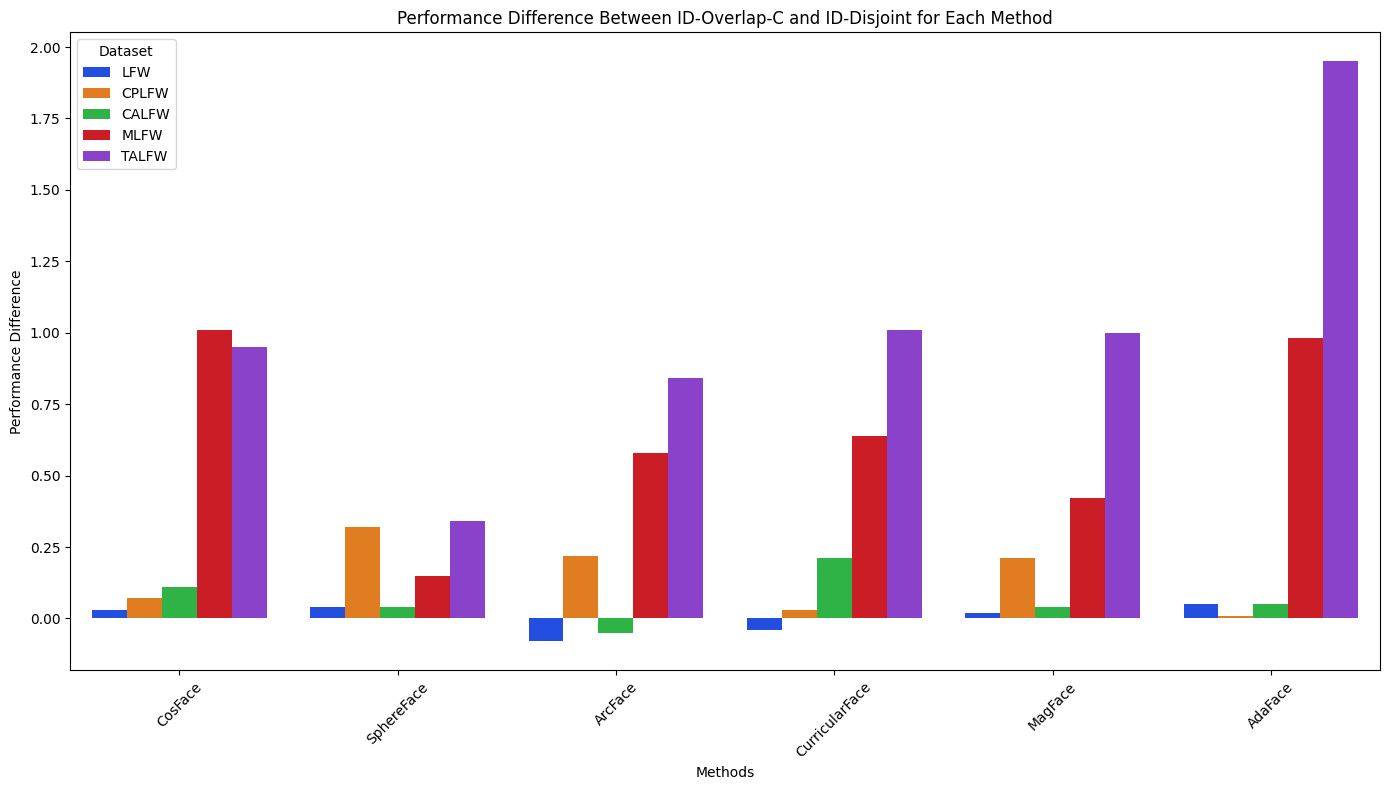}
    \end{subfigure}
   \caption{Importance of identity disjoint between training and test sets, where the importance is measured by the difference between the model trained with ID-Overlap-C and ID-Disjoint.}
\label{fig:importance-figure}
\end{figure*}

%% file: sec/6_conclusion.tex
\section{Conclusions}
\label{sec:conclusion}

We first show that 46.93\% of identities in the LFW are also in the MS1MV2, where there are 384 same images in both datasets. Based on the similarity distribution of top-two matched pairs, it is possible to detect most of overlapped identities automatically by setting a proper threshold value. 

To better compare the effect of "seeing" identities (identity overlap) in training set, we build a platform that allows us to train face recognition methods with the same training and test configurations. Since the overlapped LFW identities do not match the same number of identities in MS1MV2, we trained three constant margin algorithms and three adaptive margin algorithms with three versions of MS1MV2 - 1) ID-Disjoint: all the overlapped identities are dropped from MS1MV2, 2) ID-Overlap-R: keeping the overlapped identities and dropping the same number of randomly selected identities, and 3) ID-Overlap-C: Same as ID-Overlap-R, but reduce the identity noise by merging the detected identity mis-categorized folders. The results show that "seeing" the test identities in the training set brings a consistent benefit to face recognition models in terms of accuracy compared to not "seeing" on accuracy. With the benefit of less identity noise, five out of six models have performed better than when trained with vanilla MS1MV2, even though the number of identities and images is smaller.

We also reveal the relationship between the effect of identity overlap and difficulty level of test sets. The results indicate that the benefit the model gets from the identity overlap increases dramatically when the difficulty level of test sets increases. Note that the smallest improvement due to the identity overlap on TALFW is twice as large as the improvement due to the algorithm.

This paper shows that identity-disjoint training and test sets is non-trivial to engineer after the datasets are independently assembled without a common set of identity labels. 
It also shows that identity-disjoint training and test sets are essential in order to avoid optimistic accuracy bias.